# Flexible Multi-Generator Model with Fused Spatiotemporal Graph for Trajectory Prediction

***Peiyuan Zhu*[1]*, Fengxia Han*[2]*, Hao Deng*[2*]**

[1]*School of Electronic and Information Engineering, Tongji University, Shanghai, China*
[2]*School of Software Engineering, Tongji University, Shanghai, China*
**denghao1984@tongji.edu.cn*

## Abstract

Trajectory prediction plays a vital role in automotive radar systems, facilitating precise tracking and decision-making in autonomous driving. Generative adversarial networks with the ability to learn a distribution over future trajectories tend to predict out-of-distribution samples, which typically occurs when the distribution of forthcoming paths comprises a blend of various manifolds that may be disconnected. To address this issue, we propose a trajectory prediction framework, which can capture the social interaction variations and model disconnected manifolds of pedestrian trajectories. Our framework is based on a fused spatiotemporal graph to better model the complex interactions of pedestrians in a scene, and a multi-generator architecture that incorporates a flexible generator selector network on generated trajectories to learn a distribution over multiple generators. We show that our framework achieves state-of-the-art performance compared with several baselines on different challenging datasets.

## 1 Introduction

Trajectory prediction is crucial for accurately predicting the behavior of pedestrians, bicycles, or vehicles in AI-based radar systems, especially in automotive radar systems. After obtaining the target's data from the automotive radar, the target's historical motion trajectory and behavior patterns can be analyzed to better predict their future actions, thus making more accurate decisions and planning for safer and more efficient autonomous driving [1].

Humans in the trajectory prediction problem tend to have complex interactions. For pedestrians in a certain traffic scene, their movements are not independent of each other. They react and adjust according to their own strategies [2]. For example, a pedestrian will adjust his future path when encountering obstacles or other pedestrians. The former generally originates from the information in the visual background, and the latter refers to social interaction between pedestrians. As there is an abundant distribution of potential human behaviors, each pedestrian may take numerous conceivable trajectories, which is known as multimodal. Imagine a pedestrian arriving at an intersection, he is likely to go straight, turn left or right. It is a fundamental truth that any two regions guided by different directions are inherently disconnected. If the pedestrian ventures towards an area that bridges these disparate directions, it means he treads upon trajectories that defy the realm of possibility, which is referred to as out-of-distribution (OOD) trajectories. In other words, the pedestrian trajectory data exhibits a multimodal property where it falls on a union of several disjoint manifolds.

Recent work leverages generative adversarial networks (GANs) [1], [3], [4] to learn a distribution over pedestrian trajectories, most of which overlooks the multimodal property of pedestrian trajectories and leverages single-generator GANs. However, single-generator GANs fail to learn the disconnected multimodal distributions [5], [6], thus generating unexpected OOD samples in between manifolds. This drawback could be solved by using a collection of generators. Besides, Prior methods are mostly based on the tracking coordinate data that varies in time or the static background images. They overlook the complex variations within the interaction relationships of pedestrians along with the highly changing visual scenes. Part of the reason is that the existing pedestrian trajectory prediction datasets [7], [8], [9] merely offer coordinate information for tracking each object in the original video. It falls short, however, in providing supplementary data that allows for the identification of specific pedestrians in the video corresponding to their annotated pedestrian ID.

To address these issues, we propose a multi-generator approach with a novel interactive scheme based on the local perspective of pedestrians, which can effectively model various disconnected manifolds in pedestrian trajectory prediction scenarios. Intuitively, the impact of the surrounding environment and other pedestrians within the limited perspective of the target pedestrian is more substantial than that in a distant area. Inspired by the spatiotemporal prediction task of radar echo forecasting [10], we recombine the original trajectories and background images from the perspective of the target pedestrian into a time-varying structured graph, which represents the complex interaction between pedestrians and physical scenes and among pedestrians themselves. Then we can capture the transient variations of the structured graph over time using a spatiotemporal encoder. With the extracted highly fused interaction features, it helps to generate trajectories that are feasible physically and socially. We introduce a network named Generator Selector to learn the priors over multiple generators, thereby determining which generators are more likely to generate trajectories. The contributions of this work are summarized as follows:

- We design a novel spatiotemporal graph to integrate the complex interaction between pedestrians and surroundings, as well as among pedestrians.
- We incorporate multiple generators in our decoding process based on the proposed spatiotemporal graphs to model disconnected manifolds.
- We conduct extensive experiments to verify the effectiveness and generalization ability of our model. The experiment results demonstrate that our approach achieves state-of-the-art performance.

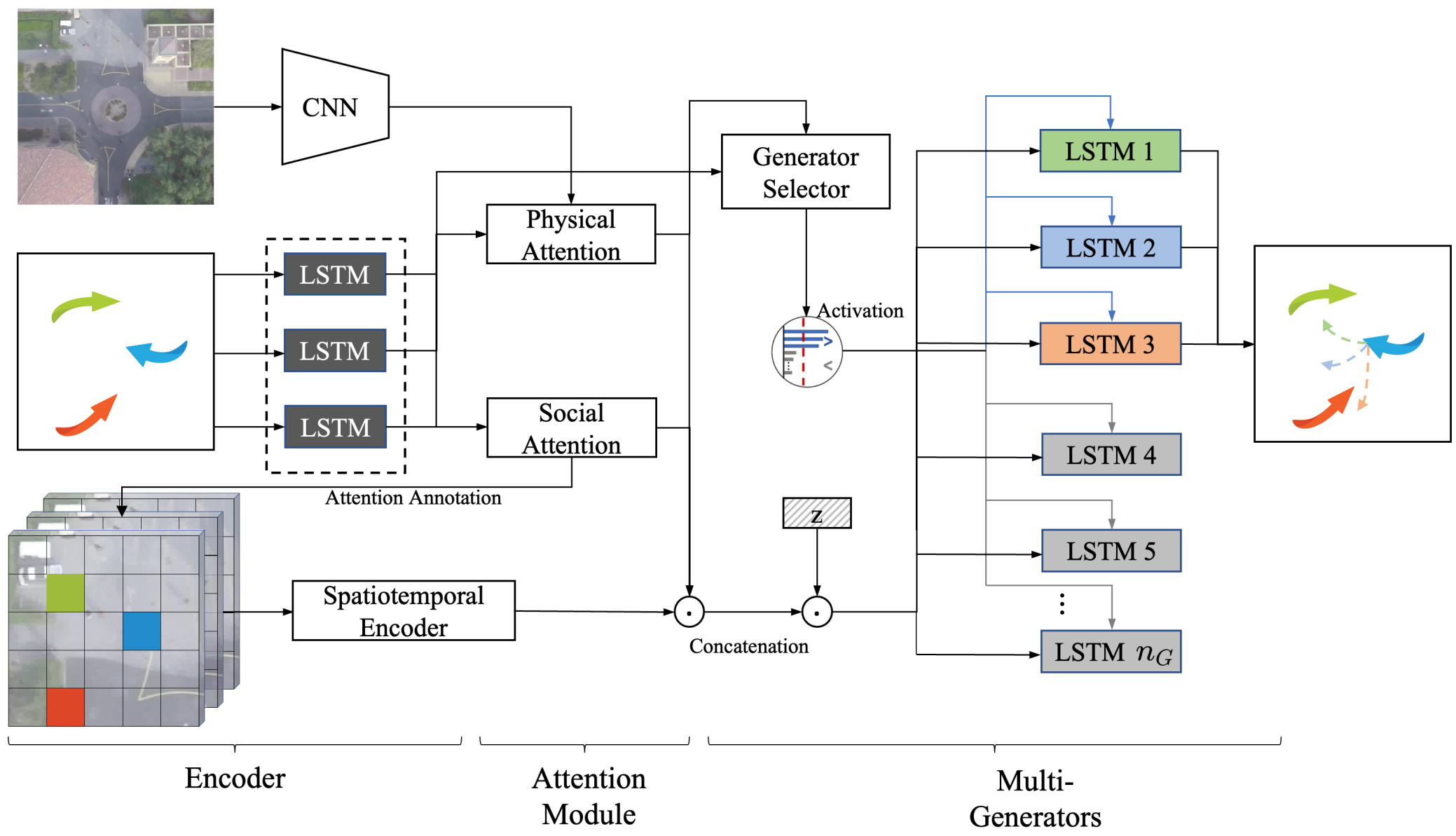

Figure 1. An overview of our model: It takes pedestrian trajectories and scene images as inputs and outputs future trajectories in disconnected manifolds.

# 2 Method

### *2.1 Problem Definition*

Pedestrian trajectory prediction is the problem of forecasting the future navigation movements of pedestrians in a given environment, which involves analyzing various factors such as the pedestrian's current position and speed, as well as situational factors like the layout of the environment and the presence of obstacles or other pedestrians. In this work, we aim to learn the multimodal trajectory distribution making full use of spatiotemporal information that can generate plausible future trajectories for each pedestrian. For a given prediction scenario, the input of the model is composed of two parts: (1) Observed pedestrian trajectory coordinate data $X_i = \{(x_i^t, y_i^t) \in \mathbb{R}^2 | t \in [t_1, t_{obs}]\}$ for each pedestrian $i$ in the previous frames. (2) The image $I$ of the scene. The model is capable of predicting a set of $K$ diverse trajectories: $\hat{Y}_i^k = \{(x_i^t, y_i^t) \in \mathbb{R}^2 | t \in [t_{obs+1}, t_{pred}], k \in [1, K]\}$.

### *2.2 Overall Model*

Figure 1 illustrates the architecture of our model. The model uses a physical encoder to extract visual features and a social encoder to learn from observed human trajectories. The observed trajectories and scene images are encoded and forwarded to the attention module. The attention module consists of two components: physical attention and social attention. The social attention is utilized to construct a sequence of structured graphs by annotating attention vectors, which is then encoded by a spatiotemporal encoder to capture transient variations of the background scene and pedestrian movement trend. The model uses the multi-generator GAN, where each generator accepts features from the concatenation of social attention, physical attention, and spatiotemporal encodings from the sequence of graphs. The Generator Selector plays a crucial role in learning the priors associated with generators, which allows us to choose a more appropriate generator using the learned priors, thereby facilitating the generation of future trajectories. The generators that exceed the set threshold will be activated and specialized to generate trajectories of a unique and disconnected manifold conditioned on the concatenated features and the noise vector.

### *2.3 Fused Spatiotemporal Graph*

Based on the idea that predicting pedestrian trajectories requires considering both the surrounding pedestrians and scenes together within the target pedestrian's perspective, we first encode the scene images and pedestrian trajectories and calculate attention, then we extract interactive features during the construction of the graphs and the spatiotemporal encoding of the graph sequence.

In this paper, we encode the visual image and observation trajectories into high-dimensional representations separately by a physical encoder and a social encoder. The physical encoder utilizes Convolutional Neural Networks (CNN) to extract visual features $V_p$ from the scene image *I*. The social encoder takes the $X_i$ sequence as input to learn the representation of observed pedestrian trajectories. To achieve this, we leverage a multilayer perceptron (MLP) to map the $X_i$ sequence into a higher-dimensional space. Subsequently, an LSTM is utilized to encode the embedding sequence, generating the representation $V_{en}^t(i)$ for pedestrian *i* at time *t*:

$$V_p = CNN(I; W_{cnn}) \tag{1}$$

$$V_{en}^t(i) = LSTM_{en}\left(MLP_{emb}\left(X_i^t, W_{emb}\right), h_{en}^t(i); W_{en}\right) \tag{2}$$

To obtain the context vector $C_{ph}(i)$ for physical attention, we employ soft attention with inputs visual feature $V_p$ extracted from the scene image and $V_{en}^t(i)$ obtained from the LSTM encoder. For social attention, we first compute the attention scores between neighboring pedestrians, taking into account the distance and bearing angle. These scores are then used to weight the embedding $V_{en}^t(i)$ of the encoder LSTM to obtain social features $V_{so}^t(i)$ . Following the multi-generator architecture, the generator prior is derived through the Generator Selector, and the hidden states at time $t_{obs}$ for pedestrian $i$ from each generator are weighted to obtain $\bar{h}_{dec}(i)$ from the prior. $\bar{h}_{dec}(i)$ contains abundant information for predicting the future path of the pedestrian, which we take as input along with $V_{so}^t(i)$ to attain the social attention vector $C_{so}^t(i)$ . The physical attention network $ATT_p$ and the social attention network $ATT_s$ are parameterized by $W_{ph}$ and $W_{so}$, respectively:

$$C_{ph}^t(i) = ATT_p\left(V_p, V_{en}^t(i); W_{ph}\right), \tag{3}$$

$$C_{so}^t(i) = ATT_s\left(V_{so}^t(i), \bar{h}_{dec}^t(i); W_{so}\right). \tag{4}$$

The construction of the fused spatiotemporal graph is enlightened by the characteristics of radar echo frames. Radar echo frames serve as invaluable tools in depicting the patterns of aggregated precipitation bands. Within these frames, distinct locations correspond to diverse intensities, offering opportunities for in-depth analysis of echo shapes and structural organization to ascertain the extent, morphology, and evolving trends of precipitation bands. Consequently, such insights foster accurate predictions concerning the future trajectory and impact areas of these precipitation patterns.

To construct the graph, we first crop a local image of size *H* ×*W* obtained from the global image *I* centered on the target pedestrian. In order to give the graph clear semantics, the global image can be a semantic map or an occupation map. Next, we construct a graph that is the same size as the cropped image with an additional channel dimension and then partition it into a 2D grid. We encode each ground-truth trajectory coordinate $X_t$ into the nearest grid cell representing where the pedestrian is at time *t*. We concatenate the observation encoding $V_{en}^t(i)$ and social attention vector $C_{so}^t(i)$ to obtain the annotation vector for pedestrian i. In all the grid cells that contain pedestrians, the annotation vector of the target pedestrian and other pedestrians in the local area are filled in the channel dimension, which is what we call attention annotation. The construction of our structured spatiotemporal graph is shown in Fig. 2. The sequence composed of spatiotemporal graphs annotated with time will be encoded by a spatiotemporal encoder. Note that we leverage an established spatiotemporal encoder [10], as our contribution is the well-structured graph sequence. The integrated contextual effects of incorporating surrounding pedestrians and the environment in the local perspective of pedestrians in spatiotemporal graphs are analogous to the echo intensity in radar echo frames. The variations in these effects, as features, can provide more detailed information for predicting future trajectories. During the construction and spatiotemporal encoding of the graph sequence, the interaction features are highly fused, enabling the model to capture the underlying patterns in pedestrians' movements over time.

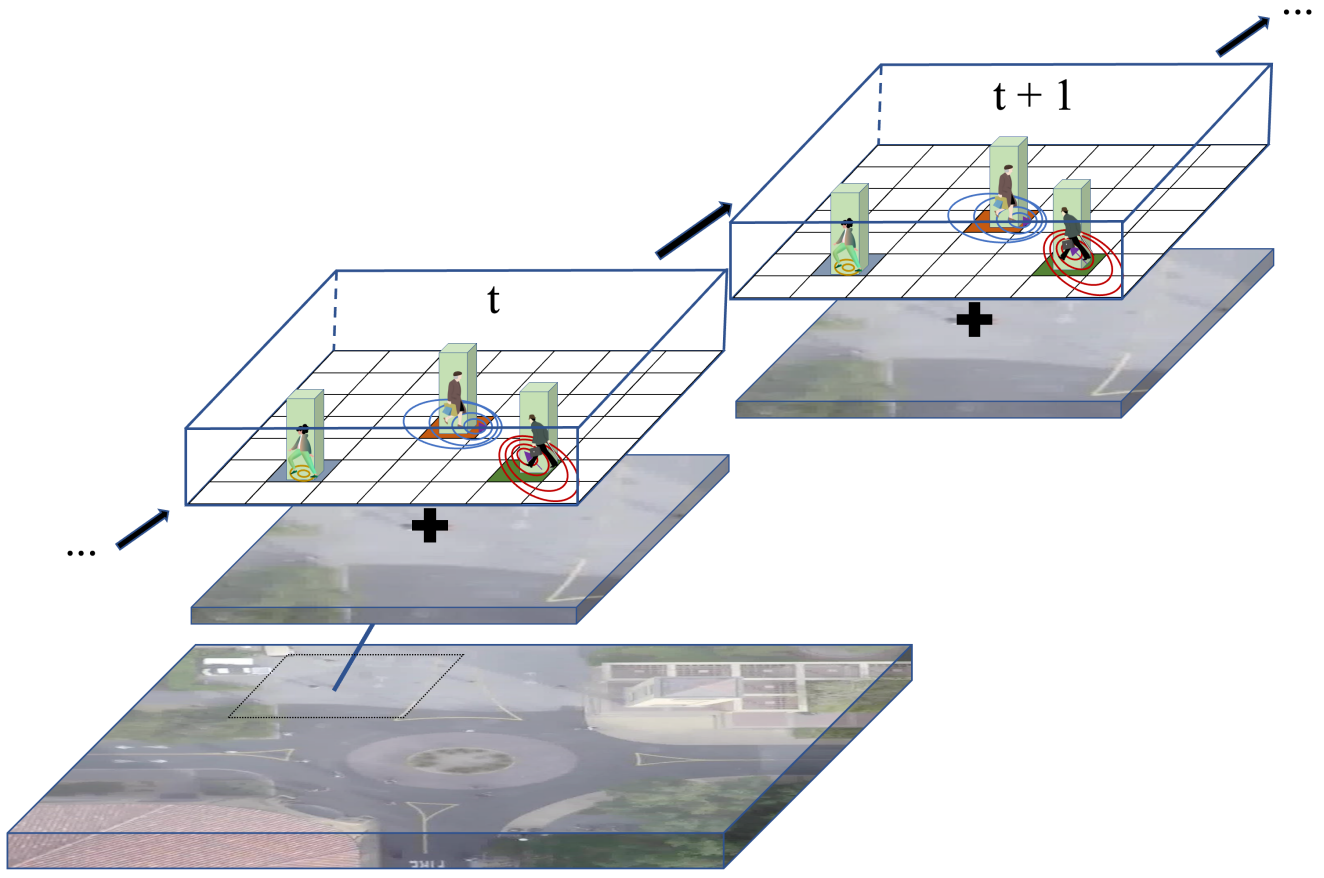

Figure 2. The construction of our structured spatiotemporal graph.

### *2.4 Generator Selector*

Our multi-generator model consists of multiple generators, a discriminator, and a classifier. We construct $n_G$ generators, where all generators share the same network architecture without sharing weights. Each generator is composed of an LSTM decoder which takes in the pedestrian trajectory encoding $V_{en}(i)$ , the physical scene context $C_{ph}(i)$ , the social scene context $C_{so}(i)$, the spatiotemporal context $C_{st}(i)$ and a random noise vector *z* sampled from a normal distribution. These features are all concatenated together as $C^{gen}(i) = [V_{en}(i), C_{ph}(i), C_{so}(i), C_{st}(i)]$. Thus the future trajectory for pedestrian *i* is obtained by:

$$\hat{Y}_i = MLP_{gen}(LSTM_g([C^{gen}(i), z], h_{dec}(i); W_{dec}); W_{gen}), \tag{5}$$

where $g \in [1, n_G]$ represents the index of generators.

The discriminator needs to additionally consider the ground-truth trajectory $Y$ and the generated trajectory $\hat{Y}$. After going through a similar process as how generators obtain $C^{gen}(i)$ , the discriminator obtains $C^{dis}(i)$ . We compute the classification score using an MLP:

$$\hat{L}_i^{dis} = MLP_{dis}\left(C^{dis}(i); W_{dis}\right). \tag{6}$$

The classifier stimulates the generators to cover the disconnected mixture manifolds of trajectories and identify which generator generates the trajectory. In our model, the classifier takes in $C^{dis}(i)$ as well and computes the classification scores using an MLP similar to the one in the discriminator:

$$\hat{L}_i^{cls} = MLP_{cls}\left(C^{dis}(i); W_{cls}\right), \quad (7)$$

where $\hat{L}_i^{cls}$ is a $n_G$- class classification score that represents the probability of the generated trajectory belonging to each generator.

In real-world situations, it is common for pedestrians to have diverse inclinations toward walking in different directions, implying that trajectories can be generated by various generators with differing probabilities. The need to understand the preferred walking direction of pedestrians or the generator that is more likely to generate them has motivated us to propose the network named Generator Selector, which is constructed using MLPs:

$$\hat{L}_g^{sel} = MLP_{sel}(C^{gen}(i); W_{sel}). \quad (8)$$

Here, $g \in [1, n_G]$ represents the index of the generator, and $\hat{L}_g^{sel}$ represents the prior assigned to each generator.

*2.5 Model Training*

We hope to derive the optimal prior $p(g)$ over the generators based on the attention-encoded feature $C^{gen}(i)$ and the predicted trajectory $\hat{Y}$. We use $q(g|Y)$ to represent the probability of future trajectory $Y$ belonging to each generator $G_g$, and approximate the true posterior distribution $p(g|Y)$ with $q(g|Y)$. We train the Generator Selector and generators alternatively. When we finish the step of training the Generator Selector, the expected value of $q(g|Y)$ over the real data distribution provides us with an approximation of $p(g)$, which we can use to train the generators.

*2.5.1 GAN Training:* Our GAN training loss function consists of three parts:

- Adversarial loss $\mathcal{L}_{adv}$, for which we use the original adversarial loss. Considering the distribution of $n_G$ generators, the calculations of each generator need to be weighted: $\sum_{g=1}^{n_G} \zeta_g p_{G_g}(\hat{Y} \mid C^{gen}(i))$, where $\zeta_g$ denotes the probability of the generator with index $g$ in the parameterized distribution $s(g;\zeta)$ obtained from the Generator Selector.
- Trajectory diversity loss $\mathcal{L}_{variety}$, which encourages each generator to produce diverse trajectories.
- Classification loss $\mathcal{L}_{cls}$, for which we use cross-entropy to calculate the classification loss, which measures between the generator index output by the classifier and the true generator index responsible for generating the trajectory.

Overall, the training objective can be summarized as follows:

$$\min_{G} \max_{C,D} \mathcal{L}_{adv} + \lambda_{variety}\mathcal{L}_{variety} + \lambda_{cls}L_{cls}, \quad (9)$$

where $\lambda_{variety}$ and $\lambda_{cls}$ are hyperparameters.

*2.5.2 Generator Selector Training:* The purpose is to learn the prior probability distribution $s(g;\zeta)$ of each generator to approximate the optimal prior $p(g)$. In order to estimate which generator is more likely to generate the predicted trajectory, inspired by [6], we use the Monte Carlo simulation to sample multiple trajectories. We approximate $p(Y|g)$ using

$$\frac{1}{l}\sum_{i=1}^{l} \mathcal{N}(Y; \hat{Y}_{z(i),g}, \sigma I). \quad (10)$$

$\mathcal{N}$ is a Gaussian distribution with variance $\sigma$, $l$ is the number of Monte Carlo samples. By applying Bayes' theorem, we obtain the conditional probability of each generator:

$$p(g|Y) = \frac{p(Y|g)}{\sum_{g'}^{n_G} p(Y|g')}. \quad (11)$$

Finally, we optimize the Generator Selector (S) by minimizing the cross-entropy loss:

$$\min_{S} H(p(g|Y), s(g)). \quad (12)$$

# 3 EXPERIMENTS

*3.1 Experimental Setup*

We evaluate on three publicly available real-world datasets: ETH [7], UCY [8], and SDD [9], and one simulated dataset: Forking Paths Dataset [11]. ETH and UCY datasets contain richly annotated trajectories of 1536 socially interacting pedestrians in five real-world scenes: ETH, Hotel, Univ, Zara01, and Zara02. SDD consists of trajectory information from approximately 20,000 objects gathered from eight distinct scenes on the campus. The Forking Paths Dataset is a multi-view dataset constructed by simulating pedestrian behavior in the real world through manual decision-making, which offers various future trajectories given a single input. We also simulate multiple future trajectories for each pedestrian on the SDD dataset to verify the model's ability to learn multiple disconnected manifolds, as well as the flexible control of generator priors.

We utilize average displacement error (ADE) and final displacement error (FDE) to evaluate the accuracy of predictions. ADE measures the average L2 distance between the ground truth and our prediction over the entire predicted time steps. FDE measures the distance between the predicted final destination and the true final destination at the end of the prediction period $t_{pred}$. The strategy of selecting the minimum among $K$ trajectories [3] is utilized, where $K$ is set to 20. We also follow the method [6] to calculate precision and recall to evaluate the ability to reduce OOD samples in trajectory prediction. Precision evaluates the percentage of generated trajectories that are correctly identified as non-OOD samples, while recall measures the extent to which the generated trajectories cover the entire range of potential trajectories in the ground truth. For precision and recall, higher values indicate better results, while lower results are preferred for ADE and FDE.

*3.2 Implementation Details*

We use Adam for training with an initial learning rate of 0.0002. The weight coefficients $\lambda_{variety}$, $\lambda_{cls}$, and the variance $\sigma$ are set to 1. We set the training samples of the Generator Selector to 1. For the construction of the spatiotemporal graph, we set the length and width to 7. The activation threshold for the generators is set to 0.03. All models are trained with 8-timestep observations ( $t_{obs} = 8$ ) and 12-timestep predictions ($t_{pred} = 20$ ).

### 3.3 Real-World Experiments

We compare our method with some state-of-the-art methods on three datasets: ETH, UCY, and SDD in Table 1. Our model performs on par with state-of-the-art methods on ADE/FDE and achieves the best performance in the ETH and HOTEL scenes. Compared with the single-generator methods: SocialGAN, SoPhie, and Social-BiGAT, our method exhibits an enhancement of ~10% in ADE/FDE. Compared with the multi-generator approach MG-GAN, we adopt a fused spatiotemporal graph to explore additional features during interaction modeling, thus achieving better performance. Especially in Zara01 and Zara02 scenes, our proposed method exhibits an improvement of over 30% in ADE/FDE.

Table 1. Baseline models compared to our method with errors reported as ADE / FDE in meters.

| | ETH | Hotel | Univ | Zara01 | Zara02 | SDD |
|---|---|---|---|---|---|---|
| SocialLSTM [2] | 1.07/2.34 | 0.79/1.72 | 0.60/1.36 | 0.42/0.95 | 0.52/1.12 | 55.64/30.04 |
| SocialGAN [3] | 0.74/1.09 | 0.46/0.98 | 0.56/1.18 | 0.33/0.67 | 0.31/0.64 | 27.23/41.44 |
| SoPhie [4] | 0.70/1.41 | 0.76/1.67 | 0.55/1.28 | 0.34/0.68 | 0.44/0.86 | 16.28/29.39 |
| Social-BiGAT [1] | 1.07/2.28 | 0.31/0.61 | 0.52/1.16 | 0.42/0.95 | 0.32/0.72 | - |
| MG-GAN [6] | 0.49/0.94 | 0.14/0.24 | 0.54/1.07 | 0.40/0.78 | 0.32/0.61 | 13.61/25.94 |
| Trajectron++ [12] | 0.54/0.94 | 0.16/0.28 | 0.28/0.56 | 0.21/0.42 | 0.16/0.32 | 10.50/17.92 |
| AgentFormer [13] | 0.45/0.76 | 0.16/0.24 | 0.25/0.47 | 0.26/0.50 | 0.14/0.26 | - |
| Ours | 0.44/**0.67** | **0.14/0.22** | 0.46/0.83 | 0.25/0.52 | 0.19/0.37 | 11.52/19.60 |

Table 2. Results on Forking Paths Dataset.

| | Precision ↑ | Recall ↑ | F1 ↑ |
|---|---|---|---|
| Trajectron++ [12] | 0.38 | 0.96 | 0.54 |
| AgentFormer [13] | 0.41 | 0.93 | 0.57 |
| MG-GAN [6] | 0.71 | 0.89 | 0.79 |
| Ours | 0.70 | 0.91 | **0.79** |

To evaluate the model's ability to reduce OOD samples, we conduct experiments on the Forking Paths Dataset using precision and recall as evaluation metrics. This dataset has multiple artificially annotated high-fidelity future trajectories for each pedestrian. From Table 2, we observe that the ability of our method to reduce OOD samples is similar to that of the state-of-the-art MG-GAN. Although we do not achieve state-of-the-art performance in terms of ADE/FDE in all the scenes, its ability to reduce OOD samples exceeds that of all the models that achieve state-of-the-art performance in ADE/FDE. This can be attributed to the fact that by reducing these L2-based metrics, there is a trade-off in terms of increased prediction variance and higher production of OOD samples.

For qualitative evaluation, we show the predicted examples of our model for 4 scenes in ETH and UCY datasets, as shown in Fig. 3. Observed trajectories appear in blue, and the ground truth is shown in red. The predicted trajectories are shown in blue for OOD samples and in orange for non-OOD samples. Each column represents one scene, and the scenes are Hotel, ETH, Univ, and Zara01 from left to right. Each example is displayed from top to bottom and includes two visualizations: the first one shows the predicted 20 trajectories, and the second one shows the heatmap generated by 3000 samples. Our model performs well in the prediction tasks for the 4 scenes and is able to avoid obstacles in all the scenes. We observe that our model can capture subtle deviations in trajectories and display the tendency of future trajectories in the heatmaps. It also captures information about the speed of pedestrian movement and generates future trajectories of different spans based on the observed dynamics of pedestrians' trajectories.

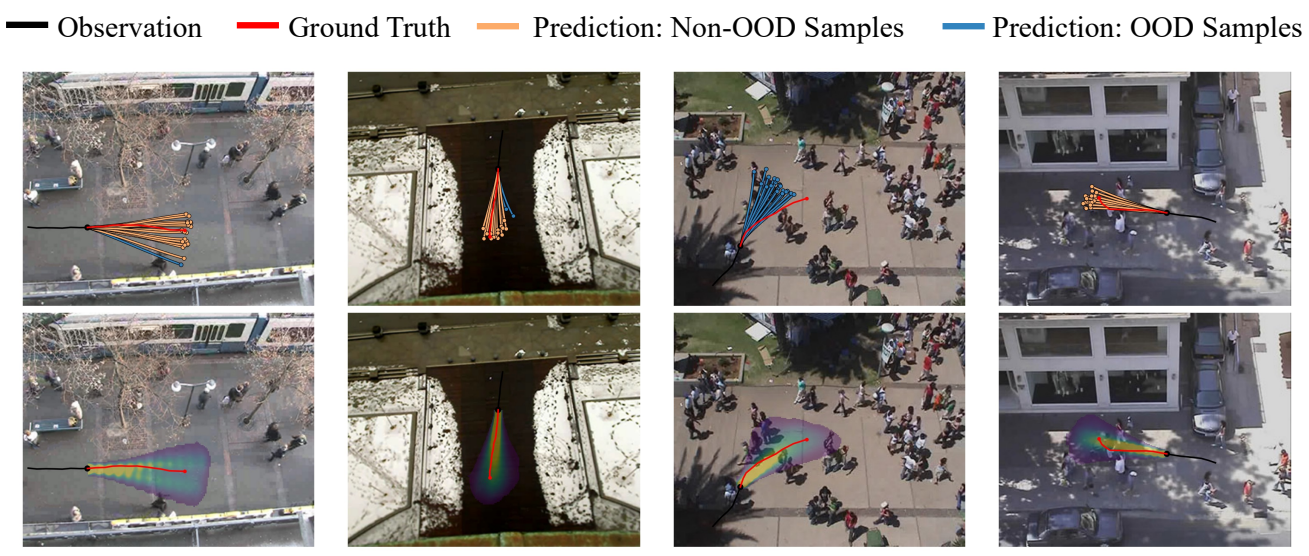

Figure 3. Qualitative results of our model on ETH/UCY.

### 3.4 Simulation Experiments

We visualize the trajectory prediction results of our model on the processed SDD dataset, as shown in Fig. 4. The figure contains two kinds of intersections for trajectory prediction. The top row shows a pedestrian passing through a T-junction, with two disconnected manifolds. The bottom row shows a pedestrian crossing a crossroad, with three disconnected manifolds. The ground truth data containing multiple manifolds is depicted in Fig. 4(a). The results of clustering the predicted trajectories, with the same color denoting the same category that should be modeled by one generator, are shown in Fig. 4(b). The prior distribution of each generator, learned by the Generator Selector, is demonstrated in Fig. 4(c), where the blue trajectories represent OOD samples.

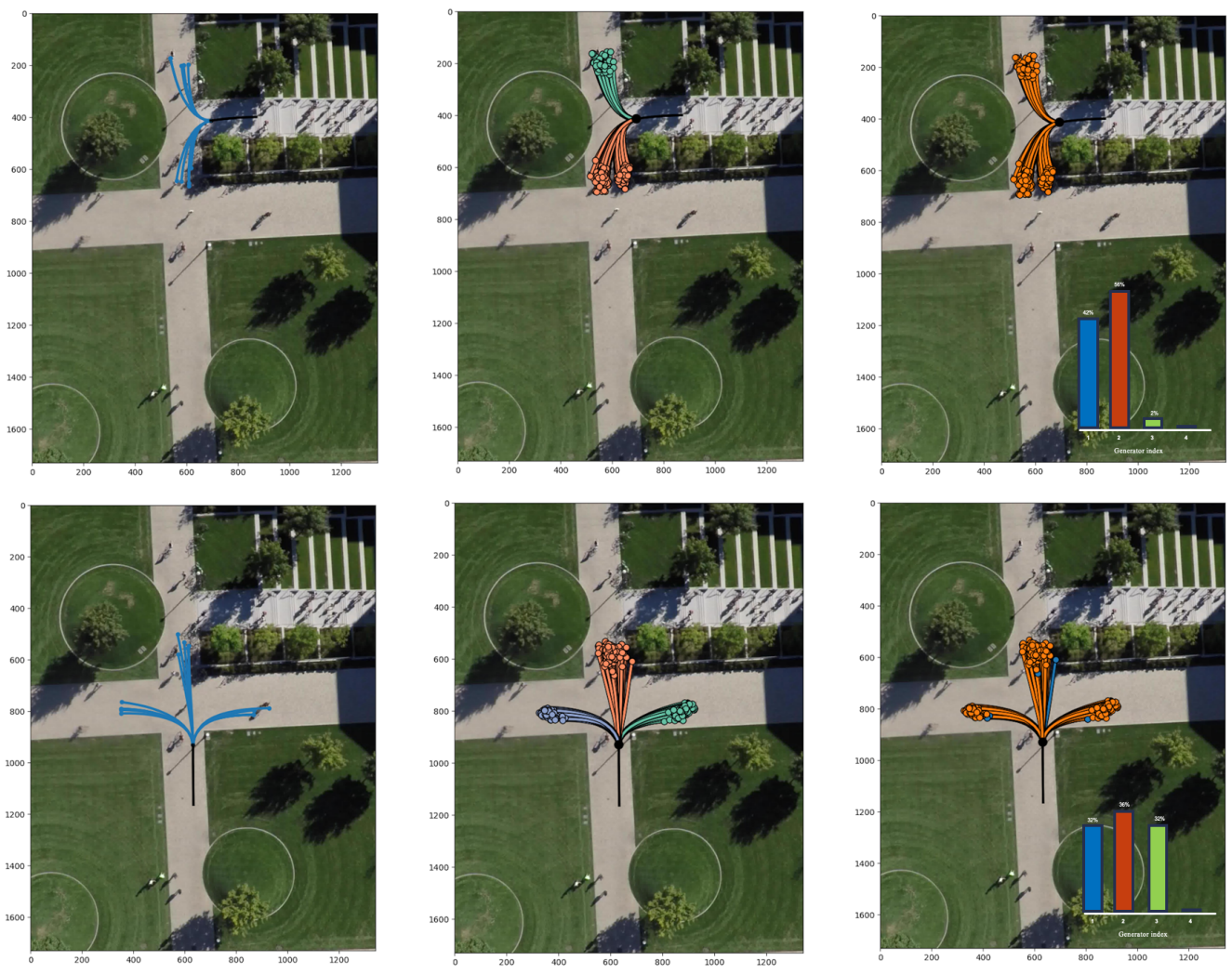

(a) Processed trajectories (b) Clustered trajectories (c) Predicted trajectories

Figure 4. Qualitative results for two kinds of intersections on SDD.

The Generator Selector learns a reasonable distribution for each generator, as shown in the bar chart in Fig. 4(c). The total number of generators is set to four, but due to the features extracted by the Generator Selector from different intersections and pedestrians, some generators may not be activated due to low probability. Specifically, two generators are not activated at the T-junction, while one is not at the crossroad. Each generator specializes in a specific manifold. By treating different intersections as distinct scenes, our model is capable of adapting to the varying scenes flexibly.

## 4 Conclusion

In this paper, we focus on the task of trajectory forecasting in automotive radar systems and learn to predict trajectories which can be seen as a mixture of multiple disconnected manifolds. Our proposed method captures the fused interaction features between pedestrians and visual scenes, as well as among neighboring pedestrians, within their localized perspective through a structured spatiotemporal graph sequence, and relies on a multi-generator architecture, where each generator specializes in learning one of the multiple disconnected manifolds in the trajectory distribution. The Generator Selector is introduced to learn the priors over multiple generators. Experimental results demonstrate the flexible adjustment capability of the prior and activation number of multiple generators, as well as its effectiveness in preventing OOD samples and the generalization ability across different datasets. We hope to demonstrate the effectiveness of the proposed structure in other domains in future work.

## 5 Acknowledgements

This work was supported in part by the National Natural Science Foundation of China under Grant 61936014, in part by the National Natural Science Foundation of China under Grant 62201388, in part by Shanghai Municipal Science and Technology Major Project No. 2021SHZDZX0100, in part by the Shanghai Science and Technology Innovation Action Plan Project 22511105300, and in part by the Fundamental Research Funds for the Central Universities under Grant 2022-5-YB-01.